\documentclass[letterpaper, 10pt, conference]{ieeeconf}

\IEEEoverridecommandlockouts
\usepackage{graphicx}
\usepackage{amsmath}
\usepackage{amssymb}
\usepackage{mathnotation}
\usepackage{cite}
\usepackage[final, defaultcolor=red]{changes}

\begin{document}

\title{Geometric Gait Optimization for Inertia-Dominated\\Systems With Nonzero Net Momentum}

\author{Yanhao Yang, Ross L. Hatton %
\thanks{This work was supported in part by NSF Grants No. 1653220, 1826446, and 1935324.}
\thanks{Y. Yang and R. L. Hatton are with the  Collaborative Robotics and Intelligent Systems (CoRIS) Institute at Oregon State University, Corvallis, OR USA. {\tt\small \{yangyanh, Ross.Hatton\}@oregonstate.edu}} %
}

\maketitle

\begin{abstract}

\looseness=-1Inertia-dominated mechanical systems can achieve net displacement by 1) periodically changing their shape (known as kinematic gait) and 2) adjusting their inertia distribution to utilize the \added{existing} nonzero net momentum (known as momentum gait). Therefore, finding the gait that most effectively utilizes the two types of locomotion in terms of the magnitude of the net momentum is a significant topic in the study of locomotion. For kinematic locomotion with zero net momentum, the geometry of optimal gaits is expressed as the equilibria of system constraint curvature flux through the surface bounded by the gait, and the cost associated with executing the gait in the metric space. In this paper, we identify the geometry of optimal gaits with nonzero net momentum effects by lifting the gait description to a time-parameterized curve in shape-time space. We also propose the variational gait optimization algorithm corresponding to the lifted geometric structure, and identify two distinct patterns in the optimal motion, determined by whether or not the kinematic and momentum gaits are concentric. The examples of systems with and without fluid-added mass demonstrate that the proposed algorithm can efficiently solve forward and turning locomotion gaits in the presence of nonzero net momentum. At any given momentum and effort limit, the proposed optimal gait that takes into account both momentum and kinematic effects outperforms the reference gaits that each only considers one of these effects.

\end{abstract}

\section{Introduction}

\begin{figure*}[!t]
\centering
\includegraphics[width=\linewidth]{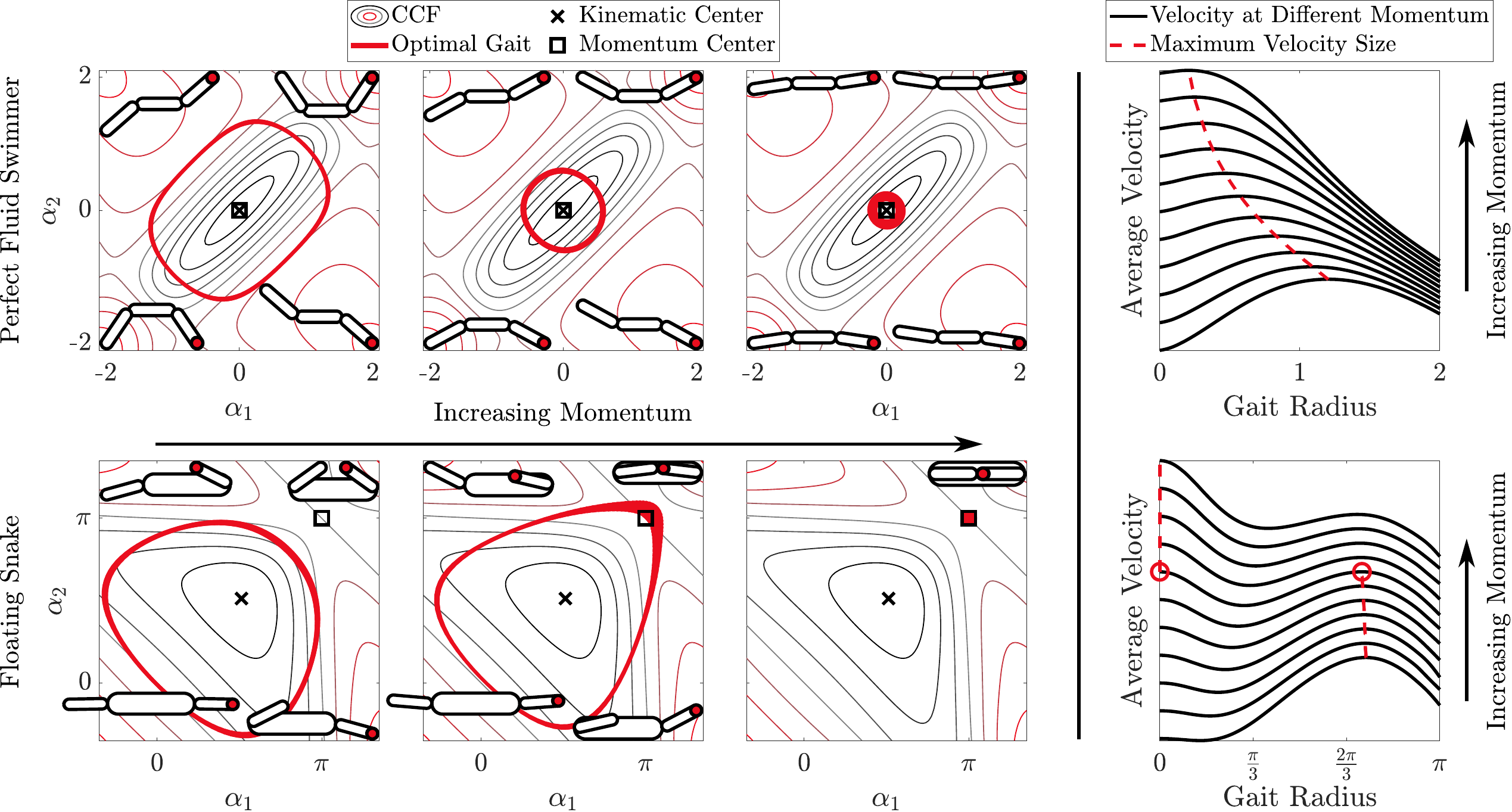}
\caption{Left: Optimal gaits of a perfect-fluid swimmer in the forward direction and a floating snake in the turning direction at different net momentum levels. \added{Inset cartoons illustrate the configuration of the system at the four different quarters of the gait, with the red markers indicating the position of the snake's or swimmer's head.} Right: Velocity and optimal gait radius of the two locomoting systems at different net momentum levels. Gaits transition from kinematic gaits to momentum gaits as the net momentum rises from zero. Depending on whether the two gaits share the same center, the transition can be continuous or discrete.}
\label{fig:intro}
\end{figure*}

Multibody systems can in many cases achieve net displacement in the absence of momentum by periodically changing their shape \cite{bloch2015nonholonomic}. Examples of such systems include falling cats \cite{montgomery1993gauge} and floating space satellites \cite{walsh1995reorienting} that are capable of reorientation without net angular momentum. Likewise, swimmers immersed in an \added{inertia-dominated} ``perfect fluid'' \added{that results in a nonzero fluid-added mass} can achieve translational and rotational self-propulsion even though the swimmer-fluid system has no net linear or angular momentum \cite{kanso2005locomotion}. If the system is assumed to start and end from rest, the net displacement induced by this effect depends only on the path of the gait through the shape space. In this paper, we will refer to gaits that maximize this displacement (relative to a time or energy cost function \cite{ramasamy2019geometry, hatton2022geometry}) as \emph{kinematic gaits}.

\looseness=-1\added{When the system has nonzero net spatial momentum resulting from prior actions, such as jumping into the air or underwater initial launch, locomotion also experiences drift.} When the shape is fixed, the drift appears as a steady translation or rotation, the magnitude of which is determined by momentum and inertia. This phenomenon occurs not only in space robotics \cite{giordano2016dynamics, nanos2017dynamics} but also in sports and dance \added{movements that are challenging for robots}, such as aerial spins in freestyle skiing and fouetté turns in ballet \cite{mcginnis2020biomechanics}. In this paper, we will refer to motions in which the system holds itself in a configuration that maximizes and directs the drift in a given direction for a given momentum as \emph{momentum gaits}.

\looseness=-1For systems with zero momentum, optimal gaits for speed or efficiency clearly require that the system actively cycle its shape (otherwise it will not translate or rotate at all). Conversely, at very large values of momentum, it is clearly best to hold a minimum-inertia shape \added{in the corresponding direction of motion}, because the kinematic contribution of any active gait will be outweighed by the loss of dynamic motion incurred by moving away from the minimum-inertia shape during this gait. Between these two extremes, however, optimal locomotion gaits tend to combine both effects -- augmenting the momentum gliding with an active pumping motion, or, equivalently, biasing kinematic gaits towards lower-inertia shapes. For example, a swimmer in a race comes out of a dive or off the wall with significant momentum, and exploits this momentum by adopting a streamlined shape that minimizes fluid-added mass in the forward direction \cite{mcginnis2020biomechanics}. As friction saps this initial momentum, swimmers transition into an active stroke that (ideally) strikes a balance between the active propulsion it supplies and the extra inertial drag it incurs \cite{lyttle2000net}.

In this paper, we examine the problem of identifying gaits that best balance kinematic and gliding contributions as a function of current momentum and available effort. 
Optimizing gaits for locomotion is highly nonlinear due to shape-dependent dynamics and the noncommutativity of translations and rotations. The geometric mechanics community has developed tools based on mathematical structures such as Lie brackets to understand these nonlinearities and to identify shape oscillations that produce useful net displacements \cite{murray1993nonholonomic, morgansen2007geometric}, and our group's previous work in coordinate optimization has further extended these tools to large-amplitude gait analysis \cite{hatton2011geometric, hatton2015nonconservativity}. 
Building on these tools, we have identified geometric structure in the optimal gaits for drag-dominated systems \cite{ramasamy2019geometry} and inertia-dominated systems with zero net momentum \cite{hatton2022geometry}, and used these insights to develop efficient algorithms for estimating the gradient of gait performance during numerical optimization. However, nonzero net momentum poses new challenges, because the original gait analysis tools cannot capture the momentum effects of the system. Considering the coupling between translations and rotations, gait optimization needs to consider both kinematic and momentum effects.

This work is also related to angular momentum control in space robotics, in which drift due to nonzero momentum is generally considered to be a disturbance. Methods for isolating this drift and compensating the control laws have been developed in \cite{giordano2016dynamics, nanos2017dynamics}. However, these methods are not designed to actively exploit the nonzero momentum. A second strain of relevant research is focused on mixed dynamic and kinematic systems such as snakeboards \cite{ostrowski1998mechanics}. Such a system can not only have nonzero momentum but can also change momentum through nonholonomic constraints. Existing works on gait design for such systems include small-time local controllability analysis \cite{ostrowski1997controllability}, optimal control \cite{ostrowski2000optimal}, and geometry-based gait selection \cite{shammas2007towards}. However, these works have not yet been extended to optimality or geometric interpretability similar to kinematic gait design. \added{Lastly, the reinforcement learning policy applied to the perfect-fluid swimmer demonstrates robustness to momentum drift \cite{jiao2021learning}. However, the drift-aware policy degenerates to suboptimal when momentum reaches zero.}

In this paper, we propose geometric gait analysis tools that consider nonzero net \added{spatial} momentum effects and the corresponding variational gait optimization algorithm. The main contributions are:
\begin{enumerate}
\item Lifting the gait description from a time-parameterized curve through the shape-space of the system (in which the shape locus determines net displacement, and the time parameterization only affects the execution cost) into a time-parameterized curve through shape-time space (which can capture the property that the pacing and duration of the gait affect the amount and nature of momentum drift over a cycle);
\item Adapting the geometric expressions for the gradient of gait performance to \deleted{work with} this lifted geometric structure; and
\item Contrasting the nature of optimal gaits for systems whose kinematic and momentum gaits have coincident or noncoincident centers.\footnote{The center of a momentum gait can be clearly defined as the configuration that minimizes inertia. The center of a kinematic gait is in some cases ambiguous. In this paper, we define the center of a kinematic gait as the point to which the gait eventually converges when \deleted{nonlinearly} increasing the cost of gait execution.}
\end{enumerate}
As a demonstration of this approach, we consider two characteristic example systems in this paper as shown in Fig.~\ref{fig:system} -- multi-link systems either immersed in an inertia-dominated ``perfect fluid'' (the \emph{perfect-fluid swimmer}) or under free-fall/zero-gravity conditions (the \emph{floating snake}), and use the proposed algorithm to determine their optimal gait for forward and turning directions, respectively. As illustrated in Fig.~\ref{fig:intro}, with the increase of net momentum, the optimal gaits transition from kinematic gaits to momentum gaits. 
Depending on whether or not the two gait centers coincide, the transition may be continuous or discrete.
The results show that, at any given momentum and effort limit, the proposed optimal gait that takes into account both momentum and kinematic effects outperforms the reference gaits that each only considers one of these effects.

\section{Background}

\added{This section reviews several key techniques of geometric mechanics used by the method proposed in this paper.}

\subsection{Inertia-dominated Mechanical System}

\begin{figure}[!t]
\centering
\includegraphics[width=0.5\linewidth]{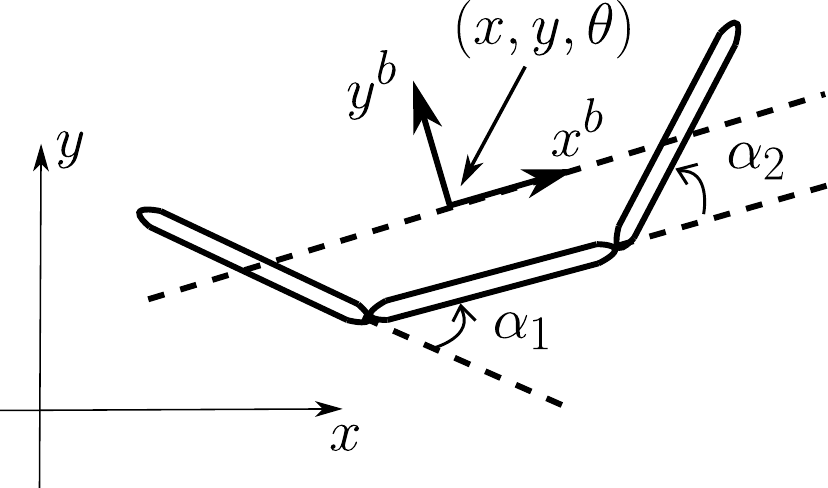}
\caption{Schematic of the kinematic system used for method validation. The system consists of three equivalent links and two joints. Depending on the medium in which it exists, the link mass can be entirely inertial or have an additional fluid mass. \deleted{For better visualization, we chose to make the center link of the floating snake twice as long as the arms, and have all the links of the perfect-fluid swimmer be the same length.} Figure reproduced from \cite{hatton2011geometric}.}
\label{fig:system}
\end{figure}

The configuration space $\bundlespace$ of a multi-body locomoting system (the space of its generalized coordinates $\bundle$) can be divided into \emph{position} space $\fiberspace$, which locates the system in the world, and \emph{shape} space $\basespace$, which gives the relative arrangement of the system components.\footnote{Formally, this decoupling is embodied in a differential geometric structure called a \emph{fiber bundle}, in which the shape and position of the system are respectively referred to as elements of the \emph{base} and \emph{fiber} spaces.} For example, the position of the three-link systems (an abstraction of perfect-fluid swimmers and floating snakes) in Fig.~\ref{fig:system} can be the center of mass and the central link orientation, $\fiber = (x,y,\theta) \in SE(2)$, and their shape \added{is the angle of the two joints}, $\base = (\alpha_{1},\alpha_{2})$.

The dynamics of an inertia-dominated locomoting system are determined by its generalized inertia matrix $M$, which relates the generalized momentum of the system to its generalized velocities,
\beqbracket \label{eq:momentum}
\begin{bmatrix} \bodymom \\ \genmom_{\base} \end{bmatrix} = \overbrace{\begin{bmatrix} M_{\fiber\fiber} & M_{\fiber\base} \\ M_{\base\fiber} & M_{\base \base}\end{bmatrix}}^{M (\base)} \begin{bmatrix} \bodyvel \\ \basedot \end{bmatrix},
\eeqbracket
where $\bodyvel=\fiber^{-1}\fiberdot$ is the body velocity of the system, $\base$ is its current shape, $\basedot$ is the shape velocity, $\bodymom$ is its generalized momentum expressed in coordinates instantaneously aligned with the body frame, and $\genmom_{\base}$ is its momentum in the shape direction.\footnote{The ``open circle'' notation we use here is similar to the ``dot'' notation, but denotes derivatives with respect to group actions rather than coordinate values. An accompanying right arrow specifies that the group operation is a right operation (locally with respect to the current configuration), and a left arrow specifies a left operation (globally with respect to the origin of the working coordinate system). The ``circle-arrow'' superscripts represent tangent vectors, such as velocity, and the subscripts are for covectors such as force or momentum. By convention, velocity is represented as the time derivative of the configuration, and momentum and force are given their own symbols. 

For example, the full reading of the notation of the body velocity $\bodyvel$ is ``the velocity constructed by taking the derivative of position with respect to right group actions'', and the full reading of the notation of spatial momentum $\spatialmom$ is ``the momentum constructed by taking the derivative of kinetic energy with respect to left (spatial) velocities''.} For a multi-body system, the inertia matrix can be constructed in the same manner as that used in \cite{hatton2022geometry}.

If the system starts with a nonzero spatial momentum $\spatialmom$ relative to the origin and there are no external forces, then the conservation of momentum means that $\spatialmom$ remains constant for all time. The top half of~\eqref{eq:momentum} can then be rewritten as
\beqbracket \label{eq:momentumconstraint}
\begin{bmatrix} M_{\fiber\fiber} & M_{\fiber\base}  \end{bmatrix} \begin{bmatrix} \bodyvel \\ \basedot \end{bmatrix} = \Adj^{*}_{\fiber} \, \spatialmom,
\eeqbracket
where $\Adj^{*}_{\fiber}$ is the dual adjoint operator that converts spatial momentum $\spatialmom$ to body momentum $\bodymom$.\footnote{In colloquial terms, $\Adj^{*}_{\fiber}$ combines the cross-product operation that converts linear and angular momentum about the origin to linear and angular momentum about the system's current location with a rotation operation that expresses the momentum in body-aligned coordinates \cite[Appendix B]{ramasamy2019geometry}.} Rearranging the momentum equation gives the system \emph{reconstruction equation},
\beq \label{eq:momentumkinrecon}
\bodyvel = -\overbrace{\inv{M}_{\fiber\fiber}M_{\fiber\base}}^{\mixedconn(\base)} \basedot + \overbrace{\inv{M}_{\fiber\fiber}\Adj^{*}_{\fiber}}^{\genmomconn(\base, \fiber)} \spatialmom,
\eeq
where the matrix $\mixedconn$ is the \emph{local connection} of the system and $\genmomconn$ is the \emph{momentum distribution function} \cite{bloch2015nonholonomic, hatton2011geometric}. Refer to the example of swim race, each row of $-\mixedconn$ encodes how the active stroke induces body velocity in each direction and can be visualized as an arrow field over the shape space, $\genmomconn(\base, \fiber) \spatialmom$ describes how much the swimmer \added{is} gliding driven by momentum, and $\genmomconn(\base, \fiber)$ itself describes how the spatially-constant momentum induces gliding in the body frame at the current configuration. Fig.~\ref{fig:local_connection} illustrates an example of body velocity reconstruction for the systems studied in this paper.

\subsection{Constraint Curvature Functions for Zero Net Momentum Systems}

Due to range-of-motion limits on the reachable shape space, locomoting systems typically move via cyclic gaits that maximize motion in a given direction relative to some execution cost. For systems with zero net momentum, the geometric mechanics community has developed methods for finding optimal gaits based on the \emph{constraint curvature} (a measure of net displacement caused by the kinematic effects of periodic shape changes) \cite{murray1993nonholonomic, walsh1995reorienting, ostrowski1998mechanics, melli2006motion, morgansen2007geometric, shammas2007geometric, avron2008geometric, hatton2013geometric, hatton2022geometry}. 

\looseness=-1The core principle of these works is that the net displacement $\gaitdisp$ over the gait cycle $\gait$ on a system described by the reconstruction equation~\eqref{eq:momentumkinrecon} with zero net momentum is the line integral of the local connection along $\gait$, so the displacement induced by a gait depends only on the gait's path in shape space. More importantly, the resulting displacement can be approximated as the surface integral of the constraint curvature $D(-\mixedconn)$ of the local connection (its total Lie bracket) over the surface $\gait_{a}$ bounded by the gait cycle,
\begin{eqalign}
\gaitdisp &= \ointctrclockwise_{\gait} -\fiber\mixedconn(\base) \\ &= \iint_{\gait_{a}} \underbrace{-\extd\mixedconn + \textstyle{\sum}\big{[}\mixedconn_{i},\mixedconn_{j>i}\big{]}}_{\text{$D(-\mixedconn)$ (total Lie bracket)}} + \text{higher-order terms} \label{eq:ccf},
\end{eqalign}
where $\extd\mixedconn$, the exterior derivative of the local connection (its generalized row-wise curl), measures the nonconservative contribution to the net displacement of the gait, and the local Lie bracket $\sum\big{[}\mixedconn_{i},\mixedconn_{j>i}\big{]}$ measures the primary noncommutative contribution. Plotting these curvature terms as scalar functions over shape space (as the contour illustrated in Fig.~\ref{fig:ccf}) reveals the effect of gaits' geometry on the motions they induce. More significantly, the constraint curvature function (CCF) $D(-\mixedconn)$ encodes the derivative of the net displacement with respect to variations in the gait.

For systems with nonzero net momentum, however, the net displacement is a function not only of the gait path but also of the duration along the gait path. This makes the CCF alone unable to accurately predict the net displacement of the gait due to the additional drift induced by momentum effects and the rotational and translational coupling between kinematic and momentum effects. In this paper, we introduce a better way to account for both effects when predicting the net displacement of the gait and performing the optimization.

\subsection{Minimum Perturbation Coordinates}

The minimum perturbation coordinates presented in \cite{hatton2011geometric} place the system’s body frame at a generalized center of mass, aligned with a generalized mean orientation of the body links. These coordinates minimize the noncommutativity of the system and enable displacement approximation of large-amplitude motions by CCFs \cite{hatton2011geometric, hatton2013geometric, hatton2015nonconservativity}. In this paper, we follow the same approach. The relationship between $\mixedconn$ and $\genmomconn$ in the original coordinates (chosen for convenience and clarity in defining the system) and the minimum perturbation coordinates used in the analysis (chosen for the accuracy of the inherent approximations) is
\beq
\bodyvel_{\text {new }}=\underbrace{\left(-\inv{\Adj_{\coordtrans}}\mixedconn + \inv{\coordtrans}\grad \coordtrans\right)}_{-\mixedconn_{\text{new}}} \basedot + \underbrace{\inv{\Adj_{\coordtrans}}\genmomconn}_{\genmomconn_{\text{new}}}\spatialmom, \label{eq:copt}
\eeq
where $\coordtrans$ is the transformation from the original body frame to the new body frame, and the adjoint operator in $\genmomconn_{\text{new}}$ is also updated to match the new body frame. In the following sections, we will use the minimum perturbation coordinates for better accuracy, but we will hide the subscripts for simplicity. Although coordinate optimization is for systems with zero net momentum, the impact of nonzero net momentum on the entire system is equivalent to the effect of momentum on the single rigid body abstraction of the system. Therefore, the minimum perturbation coordinate is still valid for systems with nonzero net momentum.

\section{Gait Optimization With Nonzero Momentum}

To extend existing geometric methods for locomotion analysis and gait optimization to systems with nonzero net momentum, we propose a gait parameterization technique that accounts for the momentum effects by including time along the gait as an extra ``shape-like” parameter. In what follows, we discuss the modifications of the local connection and CCF required to introduce the time parameterization and algorithms for gait optimization of such systems.

\subsection{Time Parameterization of Local Connections With Nonzero Net Momentum}

\begin{figure}[!t]
\centering
\includegraphics[width=\linewidth]{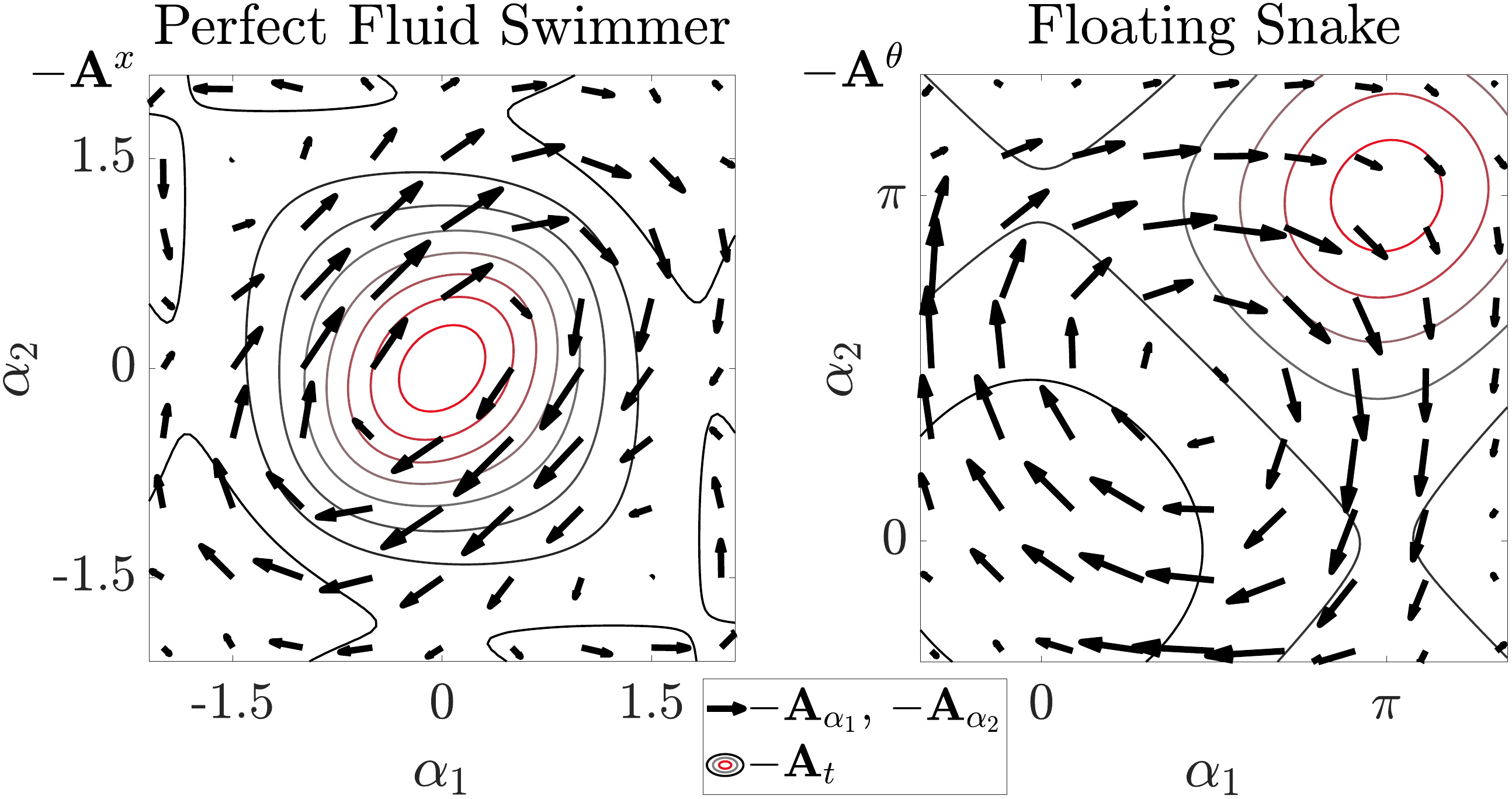}
\caption{The time parameterized local connections (or equivalently, the body velocity reconstructions) of the two locomoting systems studied in this paper in the optimized directions $x$ and $\theta$, respectively. The arrow field visualizes the local connection corresponding to kinematic motion, and the contour corresponds to the local connection of the time variable.}
\label{fig:local_connection}
\end{figure}

\begin{figure}[!t]
\centering
\includegraphics[width=\linewidth]{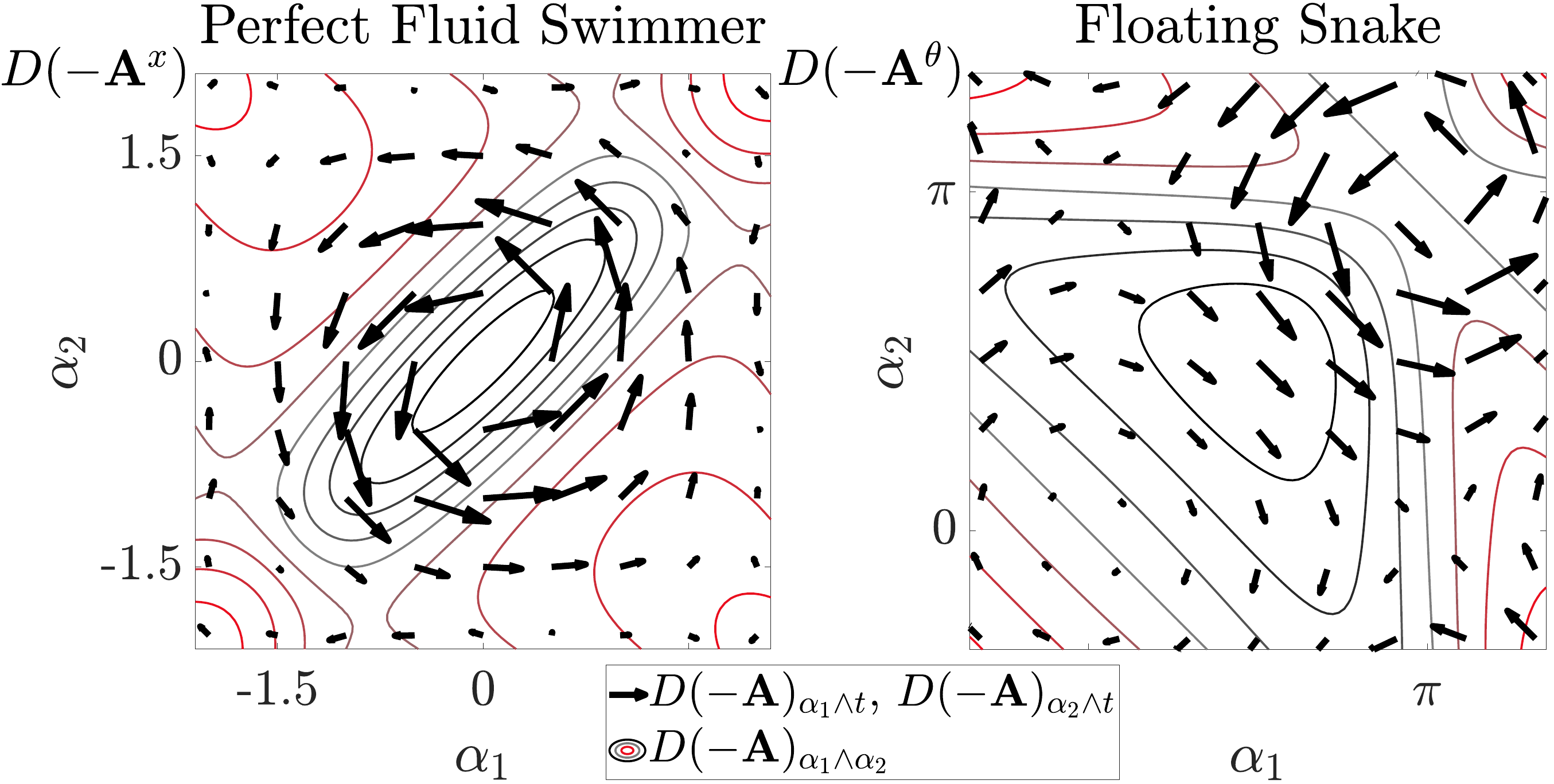}
\caption{\looseness=-1The CCFs of the two locomoting systems studied in this paper in the optimization directions $x$ and $\theta$ respectively. The arrow field visualizes the total Lie brackets between the two directions of kinematic motion and the momentum duration, and the contour corresponds to couplings within the kinematic motion. The counterclockwise arrow field means that the surface created by the gait variation towards the center can capture more CCF flux in the $d \alpha_1 \wedge d \genmomtime$ and $d \alpha_2 \wedge d \genmomtime$ directions and thus gain more momentum effect.}
\label{fig:ccf}
\end{figure}

In order to properly account for momentum effects, we construct an extended configuration space for the system in which time is treated as a system parameter along with shape and position. In our fiber-bundle partitioning of this configuration space, we include time alongside the shape parameters to form the base space, while maintaining the identification of the fiber space with the position space. Because the time variable $\genmomtime$ has the same scale as time, its derivative $\dot{\genmomtime}$ is constant with unit value. In this way, we turn the $n$-dimensional base space into $(n+1)$-dimensional base space with the constraint on the derivative of the time variable. The lifted base $\tilde{\base}$ and local connection $\tilde{\mathbf{\mixedconn}}$ become
\beq \label{eq:shapewithtime}
\tilde{\base} =\begin{bmatrix}
\alpha_1 &
\alpha_2 &
\genmomtime
\end{bmatrix}^{\top} \quad \tilde{\mixedconn}(\base, \fiber)=\begin{bmatrix}
\mixedconn(\base) & -\genmomconn(\base, \fiber) \spatialmom
\end{bmatrix},
\eeq
where $\genmomconn(\base, \fiber)$ maps the constant spatial momentum $\spatialmom$ to body velocity, and the body velocities are reconstructed as
\beq \label{eq:reconstructionwithtime}
\bodyvel = -\tilde{\mixedconn}(\base, \fiber) \dot{\tilde{\base}}.
\eeq

For a three-link system, adding the time variable makes the new local connection $\tilde{\mixedconn}$ in the reconstruction function a 3D covector field, where the magnitude of the $\genmomtime$ component is inversely proportional to the inertia of the system at the given shape and does not vary with time. Because the time variable is now part of the base, the net displacement of the system is again a function only of the lifted gait path. Note that the gait is now semi-cyclic -- the joint portion of the base variable is cyclic, yet the duration is increasing such that the start and end of the cycle are offset in the \added{$\genmomtime$} component by the gait period. 

\looseness=-1Because $\tilde{\mixedconn}(\base, \fiber)$ is not a function of $\genmomtime$, i.e. time does not affect the vector field, we can visualize this vector field in the original shape space. Fig.~\ref{fig:local_connection} shows the time-parameterized local connections in the optimized directions for the two characteristic example systems we consider in this paper. For a floating snake system, the vortex center of the kinematic motion does not overlap with the minimum-inertia configuration, whereas the opposite is true for a perfect-fluid swimmer.

For simplicity, we drop the tilde superscript in~\eqref{eq:shapewithtime} and~\eqref{eq:reconstructionwithtime} in the following sections. 

\subsection{Constraint Curvature Function With Momentum}

We can compute the system's CCF on the lifted base space in a similar manner to that used in \cite{chen2021geometric},
\beq \label{eq:tliebracketwt}
D(-\mixedconn) =&\left(-\extd \mixedconn_{1,2}+\left[\mixedconn_1, \mixedconn_2\right]\right) d \alpha_1 \wedge d \alpha_2 \\
&+\left(-\extd \mixedconn_{1,3}+\left[\mixedconn_1, \mixedconn_3\right]+\frac{\partial \mixedconn_3}{\partial \fiber} \mixedconn_1\right) d \alpha_1 \wedge d \genmomtime \\
&+\left(-\extd \mixedconn_{2,3}+\left[\mixedconn_2, \mixedconn_3\right]+\frac{\partial \mixedconn_3}{\partial \fiber} \mixedconn_2\right) d \alpha_2 \wedge d \genmomtime,
\eeq
where $\extd \mixedconn_{i, j}$ is the exterior derivative of the local connection, $\left[\mixedconn_i, \mixedconn_j\right]$ is the local Lie bracket, and the third term in the direction involving the time variable $\frac{\partial \mixedconn_3}{\partial \fiber} \mixedconn_i$ is a positional asymmetry term capturing the interaction between the position-dependent connections of the time variable because $\mixedconn_3(\base, \fiber)$ (originally $\genmomconn(\base, \fiber)$) is not only a function of shape $\base$ but also of position $\fiber$.

For a 3D base space, the CCF also has three components. Furthermore, similar to the reconstruction equation, because $\mixedconn$ is not a function of $\genmomtime$, the CCFs depend only on the joint angles. Therefore, we can again visualize the CCFs via plots on the original shape space. Specifically, in Fig.~\ref{fig:ccf}, we plot the CCFs resulting from the coupling between $\alpha_i$ and $\genmomtime$ as the arrow field, and the coupling between $\alpha_1$ and $\alpha_2$ as the contour. The visualization of CCF fits with the intuition from the time-parameterized local connections that the counterclockwise arrow field in the $d \alpha_1 \wedge d \genmomtime$ and $d \alpha_2 \wedge d \genmomtime$ directions will direct the gait variation towards minimum-inertia configuration, whereas the kinematics-related CCF in the $d \alpha_1 \wedge d \alpha_2$ direction has the richest values around its local connection vortex centers.


\subsection{Variational Gait Optimization with Nonzero Momentum}

\looseness=-1The geometry-based variational gait optimization algorithm proposed in \cite{ramasamy2019geometry} exploits the fact that the CCF flux passing through the surface enclosed by the gait is an approximation of the net displacement to compute gradients. However, the gaits studied in this paper are nonclosed in the lifted base space, which is beyond the scope of the original algorithm. 

\begin{figure}[!t]
\centering
\includegraphics[width=0.75\linewidth]{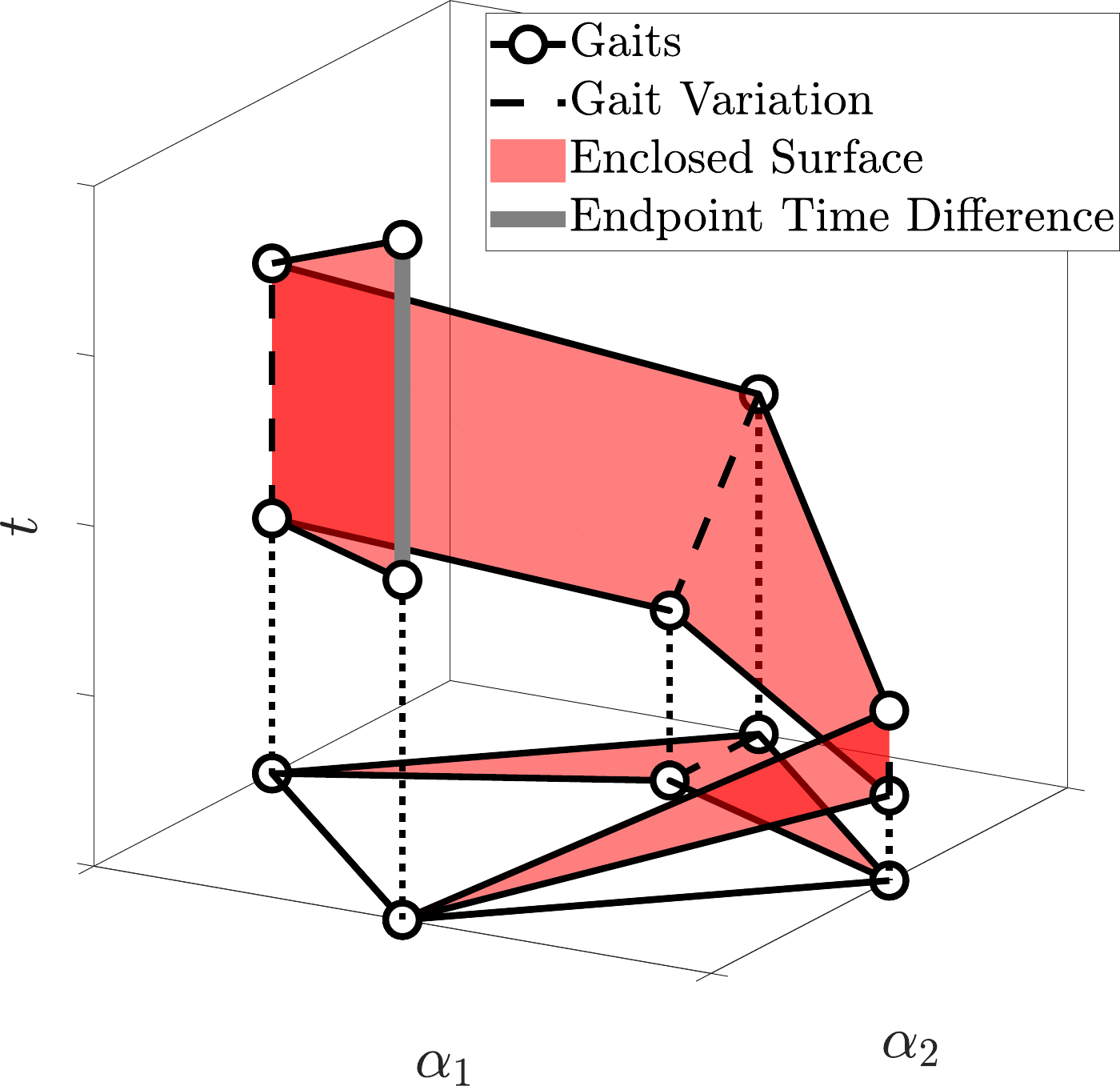}
\caption{Gait descriptions in the original 2D base space and the lifted 3D base space, and the corresponding gradient expressions for variational gait optimization for closed and nonclosed gaits. In the original 2D shape space, the gait is closed. The gradient of the net displacement with respect to the gait only involves the flux of CCF passing through the surface enclosed by the gait variation. However, the gait is no longer closed when it is lifted into a 3D base space constructed by adding the time variable (as shown by the black dotted lines). In this case, it is necessary to close the gait variation by considering the difference in the time direction at the endpoints. Thus, the gradient of the net displacement with respect to the gait contains both the flux of the CCF passing through the surface enclosed by the gait variation plus the momentum effect during the time difference at the endpoints.}
\label{fig:gait_opt}
\end{figure}

Although the net displacement cannot be approximated by calculating the area integral of the CCF now, it is still possible to form a closed surface from the gait variation by connecting the time differences at the endpoints using a method similar to that in \cite{chen2021geometric}. 
The time-difference connection that closes the gait at the endpoints actually reflects the conservative component caused by the momentum \cite{hatton2015nonconservativity}. Thus, as illustrated in Fig.~\ref{fig:gait_opt}, the complete gradient of the variational optimization algorithm for nonclosed gaits will consist of two parts: the flux of the CCF passing through the surface enclosed by the gait variation, and the momentum effect during the time difference at the endpoints. 

In this work, because we know that the gait is still closed in the joint space $(\alpha_{1},\alpha_{2})$, and the nonclosed part only involves the momentum effect duration $\genmomtime$ direction, we can directly add the displacement caused by the time difference at the gait endpoint $\phi_{\text{end}}$ to the original algorithm \added{in \cite{ramasamy2019geometry}},
\beq\label{eq:vel_grad}
\nabla_{\transpt} \fiber_{\gait} \approx \oint_\phi\added{\left(\left(\nabla_{\transpt} \phi\right)\lrcorner D(-\mixedconn)\right)} + \added{\left(\nabla_{\genmomtime}\gait_{\text{end}}\right)}\mixedconn_{\genmomtime}(\gait_{\text{end}}).
\eeq
When in 3D base space, the integral term can be calculated as moving the transcription point in the normal and binormal directions of the gait,
\beq
&\oint_\phi\left(\nabla_{\transpt} \gait\right)\lrcorner D(-\mixedconn) = \\ &\quad \oint_\phi\left(\left(\nabla_{\transpt_{\perp}} \phi\right) D(-\mixedconn)_{\| \perp}+\left(\nabla_{\transpt_{\independent}} \phi\right) D(-\mixedconn)_{\| \independent}\right),
\eeq
where the subscripts $\|$, $\perp$, and $\independent$ denote the direction tangent, normal, and binormal to the gait, and the derivatives are taken with respect to gait parameters.



To make a fair comparison between gaits with different periods, we place an upper bound $c$ on the average actuation effort $\avgenergy$ over a gait cycle \cite{hatton2022geometry},
\beq\label{eq:unit_foce_constraint}
\avgenergy = \frac{1}{T}\int_0^T\left\|\tau\right\|^2 d t\le c,
\eeq
\looseness=-1where $\lagrangeforce$ is the actuator force whose square norm gives the nonregenerable power dissipation of the actuators, and $c$ is the constant average actuation effort. In this paper, we choose $c=1$, which is the unit average actuator force constraint. 
\deleted{The kinetic energy of the system $\kineticenergy$ used to derive the Lagrangian can be calculated by substituting the reconstruction equation~\eqref{eq:momentumkinrecon} and constructing a new generalized inertia matrix $\bar{M}$ corresponding to the shape velocity and momentum}
The actuator force and its gradients with respect to gait parameters can then be calculated using the \added{similar} algorithm as we used for the momentum-free systems in \cite{hatton2022geometry}.

\added{We implement a gradient-based optimizer in MATLAB by providing~\eqref{eq:vel_grad} and~\eqref{eq:unit_foce_constraint} as the gradients and constraints for the \texttt{fmincon} optimizer using the interior-point method. We parameterize the gaits via the 4th-order Fourier series and use these series to generate direct-transcription waypoints for numerical optimization. The lower order constrains the optimizer to generate only simple gaits and prevents adjacent waypoints from intersecting each other during optimization, thus improving its numerical stability. The initial guess for gait optimization at each net momentum level is based on the optimal solution obtained at the previous level.}

\section{System Analysis}

\begin{figure}[!t]
\centering
\includegraphics[width=\linewidth]{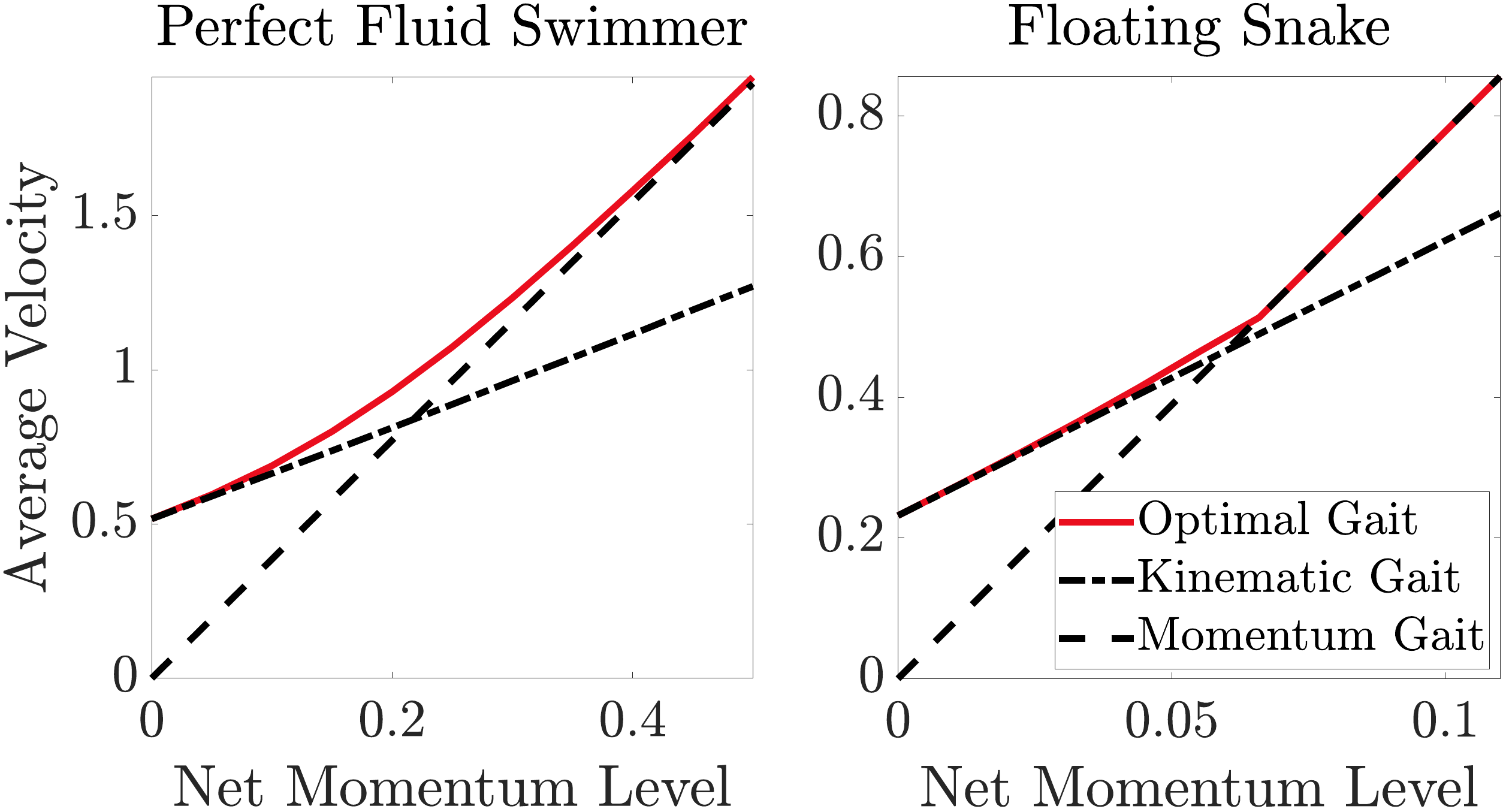}
\caption{Average forward ($x$) velocity for a perfect-fluid swimmer (left) and average angular velocity for floating snakes (right) over a gait cycle performing optimal gaits, kinematic gaits, and momentum gaits at different net forward and angular spatial momentum, respectively. All gaits are performed under the unit average actuator force constraint. The kinematic gaits are optimized without considering the contribution of momentum to displacement, and the momentum gaits are performed at the minimum-inertia configuration.}
\label{fig:velcontrib}
\end{figure}

\added{In order to evaluate the proposed method and verify the aforementioned hypotheses, we performed tests on two categories of locomoting systems with nonzero net spatial momentum $\spatialmom$}: 1) systems whose kinematic gaits have the same centers as their momentum gaits, and 2) systems whose kinematic and momentum gaits have different centers. These situations correspond respectively to the forward motion of a perfect-fluid swimmer and the turning motion of a floating snake. As illustrated in Fig.~\ref{fig:ccf}, for both of these classes of system the kinematic gaits are centered on high-value regions of the $d \alpha_1 \wedge d \alpha_2$ component of the CCF, whereas the momentum gaits maximize net displacement in the minimum-inertia configurations of the system. For different systems and directions of locomotion, these centers may or may not be aligned, leading respectively to smooth or nonsmooth transitions between these two gaits with increasing net momentum, as illustrated in the comparison of average gait velocities in Fig.~\ref{fig:velcontrib}.

\looseness=-1\added{The geometries of the two systems are shown in Fig.~\ref{fig:system}. The perfect-fluid swimmer consists of three identical links, each with a unit length, whereas the floating snake comprises a center link that is twice as long as its unit-length arms. The links are represented as ellipses with a shape aspect ratio of 0.1 and a unit density. Similarly, the fluid is also assigned a unit density. The fluid-added mass for the perfect-fluid swimmer is determined using the model proposed in \cite{hatton2013geometric}.}

\begin{figure*}[!t]
\centering
\includegraphics[width=\linewidth]{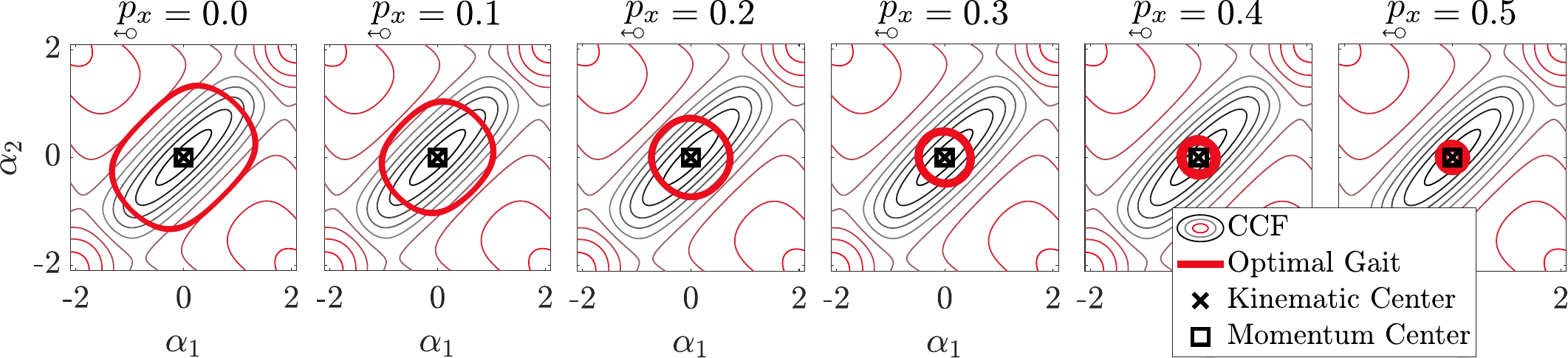}
\caption{Optimal gaits for forward ($x$) motion of a perfect-fluid swimmer constrained by unit average actuator force for different net forward spatial momentums. The pacing of the gait is represented by the line thickness, with thick lines representing ``slow'' changes in base variables and thin lines for ``fast'' changes. The contours in the background represent the CCF in the $\alpha_1\wedge\alpha_2$ direction (i.e. kinematic effects). Two markers illustrate the centers of the kinematic and the momentum gaits, respectively.}
\label{fig:gaitswimmer}
\end{figure*}

\begin{figure*}[!t]
\centering
\includegraphics[width=\linewidth]{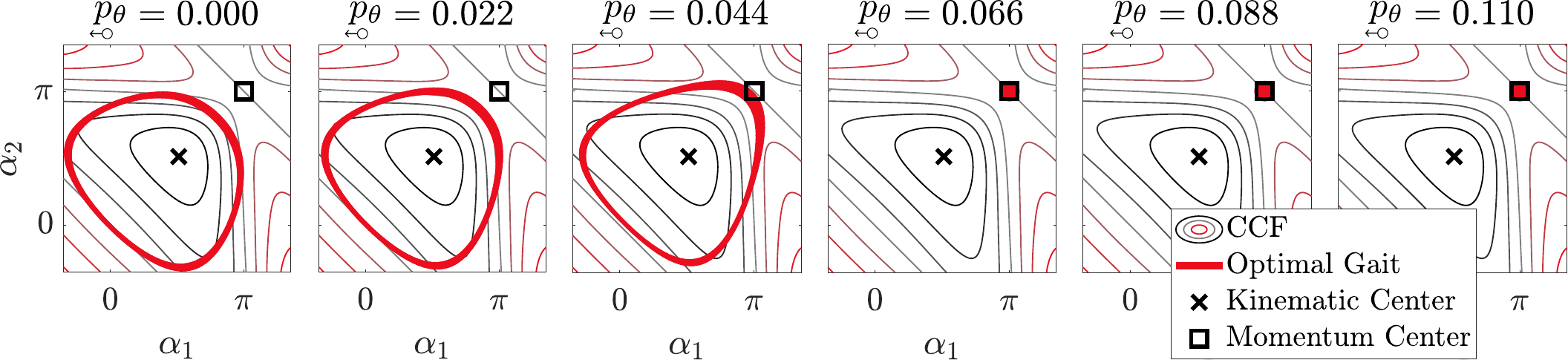}
\caption{Optimal gaits for turning motion of a floating snake constrained by unit average actuator force for different net \added{angular} spatial momentums. The pacing of the gait is represented by the line thickness, with thick lines representing ``slow'' changes in base variables and thin lines for ``fast'' changes. The contours in the background represent the CCF in the $\alpha_1\wedge\alpha_2$ direction (i.e. kinematic effects). Two markers illustrate the centers of the kinematic and the momentum gaits, respectively.}
\label{fig:gaitsnake}
\end{figure*}

\subsection{Systems With the Same Centers for Kinematic and Momentum Gaits}

An example of the kinematic and momentum gaits sharing the same center is the forward swimming of a \added{perfect-fluid swimmer that} is able to translate and rotate using cyclic shape changes \cite{kanso2005locomotion}. \deleted{The swimmer has three identical links and its geometry is shown in Fig.~\ref{fig:system}.} We compare gait optimization considering kinematics and momentum effects individually or together under the constraint of unit average actuator force.


The left panel of Fig.~\ref{fig:velcontrib} illustrates the average forward velocity for optimal, kinematic, and momentum gaits for different levels of net forward spatial momentum under the constraint of unit average actuator force. At any given momentum, the optimal gait always outperforms gaits that only consider either kinematic or momentum contributions to displacement. As net forward spatial momentum increased, the gait transitioned smoothly from fully exploiting kinematic effects to utilizing momentum. 

Fig.~\ref{fig:gaitswimmer} illustrates the optimal gaits for different net momentum levels. As mentioned before, we can clearly find that kinematic gait and momentum gait share the same center at $(0, 0)$. However, their amplitudes are different, as kinematic gaits need to cover regions rich in kinematic CCF (in $\alpha_1\wedge\alpha_2$ direction), whereas momentum gaits require to hold the system shape in the minimum-inertia configuration. As net momentum increases, momentum effects start to dominate the system. Given the unit average actuator force constraint (which prevents the system from arbitrarily increasing the speed of the gait to compensate for the increased inertia), the gait gradually transitions from a large-span kinematic gait to maintaining the minimum-inertia configuration and making more use of momentum for locomotion\footnote{\added{When the momentum reaches a sufficiently high level, the gait will ultimately converge to a point at the minimum-inertia configuration.}}.

\subsection{Systems With Different Centers for Kinematic and Momentum Gaits}

An example in which the kinematic and momentum gaits have different centers is the turning motion of a floating snake. The floating snake has no fluid-added mass, so without external force or linear momentum, it can only achieve net rotation. \deleted{The snake has a center link twice as long as its arms for better visualization.} We compare gait optimization considering kinematic and momentum effects individually or together under the constraint of unit average actuator force.


The right panel of Fig.~\ref{fig:velcontrib} illustrates the average angular velocity for optimal, kinematic, and momentum gaits for different levels of net spatial angular momentum under the constraint of unit average actuator force. At any given momentum and effort limit, the proposed optimal gait that takes into account both momentum and kinematic effects outperforms the reference gaits that each only considers one of these effects. When the net spatial angular momentum is small, the optimal gait is almost identical to the kinematic gait. Once the angular momentum exceeds this threshold, the optimal gait immediately switches to the momentum gait. 

\looseness=-1Fig.~\ref{fig:gaitsnake} illustrates the optimal gaits for different net momentum levels. \added{The kinematic gait center is located around $(1.6, 1.6)$, whereas the momentum gait center is located at $(\pi, \pi)$. Furthermore, kinematic gaits need to cover regions rich in kinematic CCF (in $\alpha_1\wedge\alpha_2$ direction), whereas momentum gaits need to maintain the minimum-inertia configuration.} As net momentum increases, the momentum effects start to increase. Given the unit average actuator force constraint, once the momentum effects outweigh the kinematic effects, the gait switches from a large-span kinematic gait to maintaining the minimum-inertia configuration and making more use of momentum for locomotion.

This discretization can be understood by considering a 1D gait optimization problem, where the gait is restricted to be a circle tangent to $(\pi, \pi)$ on the $\alpha_1=\alpha_2$ line, with variable radius and uniform pace.
Because the planar rotation is not subject to noncommutativity, we can consider the angular velocities contributed by kinematics and momentum separately. Because both the CCF around the center of the kinematic gait and the inertia around the center of the momentum gait are symmetric, they can be approximated locally as elliptic paraboloids. According to the surface integral of CCF and the proportional relationship between inertia and velocity, we can conclude that the angular velocities caused by the kinematic and momentum effects both peak at the respective centers and decrease nonlinearly away from their centers. 

\begin{figure}[!t]
\centering
\includegraphics[width=\linewidth]{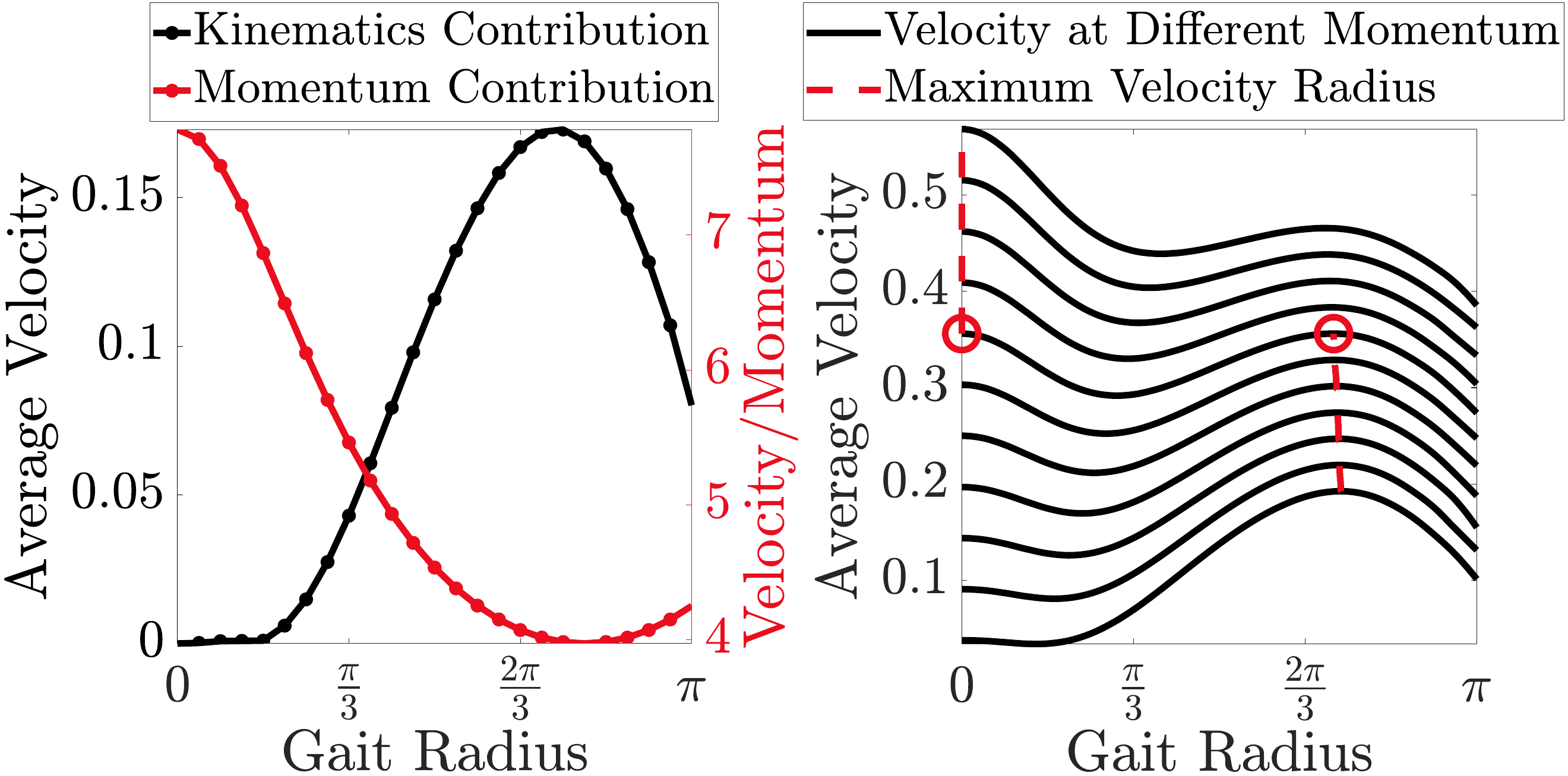}
\caption{\looseness=-1Circular gait analysis for interpreting the discrete change in optimal gaits. Left: Kinematics and momentum contributions to the average velocity for different gait radii. The momentum contribution is normalized by the momentum magnitude. Right: Total angular velocity at different net momentum levels and gait radii, and the gait radius at which maximum velocity is obtained. The positions of the two peaks of kinematic and momentum contributions do not coincide, which explains the discrete variation in optimal gaits.}
\label{fig:two_peaks_circ_gait}
\end{figure}

\looseness=-1Fig.~\ref{fig:two_peaks_circ_gait} samples these two angular velocities and \added{their sum} at different net momentum levels and gait radii. We can clearly see that there are two peaks in the total angular velocity due to the noncoincidence of \added{the center positions of the kinematic gait and the momentum gait}, and their magnitude depends on the net momentum level. Once the net momentum is greater than a certain value, the peak at the minimum-inertia configuration will exceed the one at the kinematic gait center. Therefore, as the net momentum level rises, there is a discrete switch in the optimal gait.

\section{Conclusion and Future Work}

\looseness=-1This work presents improved geometric gait analysis tools that can account for nonzero net momentum effects by lifting the gait description into time-parameterized curves in shape-time space. We also propose the variational gait optimization algorithm corresponding to the lifted geometric structure. It enables visual gait design and efficient gait optimization for inertia-dominant systems with nonzero net momentum. This approach also identifies two distinct types of systems with center-coincident and noncoincident kinematic and momentum gaits. Examples of perfect-fluid swimmers and floating snakes demonstrate that the proposed algorithm can effectively solve forward and turning motion gaits in the presence of nonzero net momentum. The optimal gaits derived by our method are consistently faster at a given effort limit than the kinematic or momentum gaits alone. 
Future work includes: \added{enabling the analysis and optimization of gait for systems with variable net momentum}; extending our methods to systems locomoting in a direction that is not aligned with its nonzero momentum, systems with mixed inertial and viscous drag, or systems in the SO(3) and SE(3) groups; adapting the Lagrange multiplier method \cite{choi2022optimal} or gait morphing algorithms \cite{deng2022enhancing} to \added{identify the optimal gait family across various net momentum levels for efficient hardware deployment; integrating the proposed method with feedback control to ensure the stability of underwater locomotion for elongated bodies presented in \cite{jing2012effects}; and enhancing the error characterization and analysis of \eqref{eq:vel_grad} in \cite{hatton2015nonconservativity, bass2022characterizing} by incorporating the impacts of nonzero net spatial momentum effects}.

\bibliographystyle{IEEEtran}
\bibliography{ref}

\end{document}